\documentclass[conference]{IEEEtran}
\IEEEoverridecommandlockouts
\usepackage{cite}
\usepackage{amsmath,amssymb,amsfonts}
\usepackage{algorithmic}
\usepackage{graphicx}
\usepackage{textcomp}
\usepackage{xcolor}
\usepackage{url}
\usepackage{multirow}
\def\BibTeX{{\rm B\kern-.05em{\sc i\kern-.025em b}\kern-.08em
    T\kern-.1667em\lower.7ex\hbox{E}\kern-.125emX}}
\begin{document}

\title{Human-Machine Interaction Speech Corpus from the ROBIN project}

\author{\IEEEauthorblockN{Vasile Păiș, Radu Ion, Andrei-Marius Avram, Elena Irimia, Verginica Barbu Mititelu, Maria Mitrofan}
\IEEEauthorblockA{\textit{Research Institute for Artificial Intelligence "Mihai Drăgănescu", Romanian Academy} \\
Bucharest, Romania \\
\{vasile,radu,andrei.avram,elena,vergi,maria\}@racai.ro}}

\maketitle

\begin{abstract}
This paper introduces a new Romanian speech corpus from the ROBIN project, called ROBIN Technical Acquisition Speech Corpus (ROBINTASC). Its main purpose was to improve the behaviour of a conversational agent, allowing human-machine interaction in the context of purchasing technical equipment. The paper contains a detailed description of the acquisition process, corpus statistics as well as an evaluation of the corpus influence on a low-latency ASR system as well as a dialogue component.
\end{abstract}


\section{Introduction}
ROBIN\footnote{\url{http://aimas.cs.pub.ro/robin/en/}} 
is a user-centred project aiming to develop software and services for human interaction with robots within a digital interconnected society. 
Its focus is on several types of robots: assistive ones - targeting users with special needs (people with some medical problems or the elderly), robots for interaction with clients and software robots that can be installed on vehicles with the aim of (semi)autonomous driving. 
One of the objectives of the ROBIN-Dialog component project\footnote{\url{http://aimas.cs.pub.ro/robin/en/robin-dialog/}} was the creation of necessary Romanian language resources and processing tools for making a robot able to communicate with users in tasks defined within several micro-worlds. One example of micro-world is given by the interaction within the notebooks department of an electronics store. This micro-world is made up of the physical space occupied by this department, by the notebooks that are commercialized by that store, their characteristics on the basis of which customers decide what products they buy, their availability, the provisional date for their becoming available, the robot and the customers who interact with the robot for finding the notebook they want to purchase or for finding the right configuration for their needs. Acting as a shop assistant, the robot must be aware of the products the department commercializes, their availability, their characteristics, as well as the types of usage scenarios they are adequate for (e.g. notebooks for gaming, for design, programming, etc.).

Previously, \cite{b1} and \cite{b2} described the natural language processing pipeline being used, as well as the dialogue manager for micro-worlds. Furthermore, \cite{b3} and \cite{b4} presented a low-latency automated speech recognition (ASR) system developed and used within the ROBIN project. This paper introduces a new speech corpus recorded for the purposes of improving the performance of the ASR system and further of the entire pipeline. The paper is structured as follows: in Section \ref{sec:related} we present related work, including other available Romanian speech resources, in Section \ref{sec:acquisition} the corpus acquisition process is described, Section \ref{sec:statistics} contains relevant corpus statistics, while in Section \ref{sec:usage} we consider the impact of the new corpus being used in the ROBIN project. We conclude the paper in Section \ref{sec:conclusions}.

\section{Related work}
\label{sec:related}

\begin{table*}[t]
    \centering
    \caption{Public Romanian speech corpora statistics.}
    \label{tab:ro_audio_datasets}
    \begin{tabular}{|l|c|c|c|c|c|}
        \hline
        \textbf{Corpus} & \textbf{Speech Type} & \textbf{Domain} & \textbf{\# Hours} & \textbf{\# Utterances} & \textbf{\# Speakers} \\
        \hline
        RSC \cite{b9} & Read & Wikipedia & 100 & 136.1k & 164 \\
        \hline
        RoDigits \cite{b10} & Read & Digits & 37.5 & 15.4k & 154 \\
        \hline
        SWARA \cite{b19} & Read & Newspapers & 21 & 19k & 17 \\
        \hline
        RO-GRID \cite{b24} & Read & General & 6.6 & 4.8k & 12 \\
        \hline
        RSS \cite{b8} & Spontaneous & Internet, TV & 5.5 & 5.7k & 3 \\
        \hline
        RASC \cite{b23} & Read & Wikipedia & 4.8k & 3k & - \\
        \hline
        CV \cite{b11} & Read & Wikipedia & 9 & 8k & 130 \\
        \hline
        VoxPopuli \cite{b15} & Spontaneous & Legal & 83 & 27k & 164 \\
        \hline 
        MaSS \cite{b15} & Read & Bible & 23 & 8.1k & 1 \\
        \hline
        \hline
        \textbf{ROBINTASC} & \textbf{Read} & \textbf{Technology} & \textbf{6.5} & \textbf{3.8k} & \textbf{6} \\
        \hline
    \end{tabular}

\end{table*}

Compared to better resourced languages, such as English, speech resources available for the Romanian language are reduced in number. The representative corpus of the contemporary Romanian language (CoRoLa)\cite{b5} contains a spoken component that can be interrogated via the OCQP platform\footnote{\url{http://corolaws.racai.ro/corola_sound_search/index.php}} \cite{b6}. Currently, it contains professional recordings from various sources (radio stations, recording studios), broadcast news and extracts from Romanian Wikipedia read by non-professionals (recorded in non-professional environments). In the context of the ReTeRom project\footnote{\url{https://www.racai.ro/p/reterom/}}, the CoBiLiRo platform\cite{b7} was built to allow gathering of additional bimodal corpora, one of the final goals being to enrich the CoRoLa corpus.

The Read Speech Corpus (RSC)\cite{b9} contains 100 hours collected from 164 native speakers, mainly students and faculty staff, with an age average of 24 years. The sentences were selected from novels, online news and from a list of words that covered all the possible syllables in Romanian.

The RoDigits\cite{b10} corpus contains 37.5 hours of spoken connected digits from 154 speakers whose ages vary between 20 and 45. Each speaker recorded 100 clips of 12 randomly generated Romanian digits, and after the semi-automated validation, the final corpus contains 15,389 audio files.

SWARA \cite{b19} is a corpus that comprises speech data collected from 17 speakers which was manually segmented at the utterance-level, resulting in a dataset of approximately 21 hours of transcribed speech, split into over 19,000 audio-text pairs.

The RO-GRID \cite{b24} dataset was developed by reading sequences of six words chosen from a list of alternatives. The first three words were designated as "keywords" and the speaker had to utter all combinations, which ended up being 400 ones. The last three words were designated as "fillers" and were randomly chosen while creating the sentence. The final corpus contained 6.6 hours of audio from 12 speakers.

The Romanian Speech Synthesis (RSS) \cite{b8} corpus was designed for speech synthesis and it contains 4 hours of speech from a single female speaker using multiple microphones. The speaker read 4,000 sentences chosen for diphtone coverage, that were extracted from novels and newspapers and fairy-tales. RSS was also extended with over 1,700 utterances from two new female speakers, comprising now 5.5 hours of speech.

Romanian Anonymous Speech Corpus (RASC) \cite{b23} is a dataset that applied the concept of crowd-sourcing to collect Romanian spoken data from the general population, by developing an open interactive platform. The corpus currently contains 4.8 hours of transcribed audio.

The Common Voice (CV) \cite{b11} corpus is a massively multilingual dataset of transcribed speech. At the moment of this writing, the Romanian version contains 9 hours of transcribed audio (6 hours validated) recorded by 130 speakers, using sentences from the Romanian Wikipedia.

VoxPopuli \cite{b15} is a large-scale multilingual corpus that contains 100,000 hours of raw audios in 23 languages and 1,800 hours of transcribed speech in 16 languages. One of the languages found in the corpus is Romanian, with 4,500 hours of unlabelled speech and 83 hours of transcribed audio.

Multilingual corpus of Sentence-aligned Spoken utterances (MaSS) \cite{b22} is a speech dataset based on readings of the Bible. The dataset contains 8,130 of parallel spoken utterances in eight languages, thus also allowing construction of end-to-end speech translation systems. The Romanian version contains 23 hours of spoken data.

Table \ref{tab:ro_audio_datasets} summarizes the statistics of the publicly available Romanian speech corpora presented above.

\section{Corpus acquisition}
\label{sec:acquisition}
The ROBINTASC corpus was collected at RACAI, during the year 2020, as part of the ROBIN project. It was recorded by a number of 6 speakers of different genders (3 males and 3 females) and ages. For recording purposes, the RELATE \cite{b12} platform was extended to allow for audio files to be stored, recorded and listened to. 

The audio processing component is activated if a corpus is created within the platform by specifying that it contains audio files. This enables all bimodal processing features. Since we start with text sentences for which we aim to provide recordings, the first step is to upload the associated texts. These can be uploaded either as separate text files or as a single CSV file containing each sentence on a different line. In the last case, the platform allows specifying the column containing the text as well as CSV characteristics such as headers, column separators, enclosing characters and optional characters indicating comments (lines to be skipped).

\begin{figure}[h]
\centerline{\includegraphics[width=0.9\linewidth]{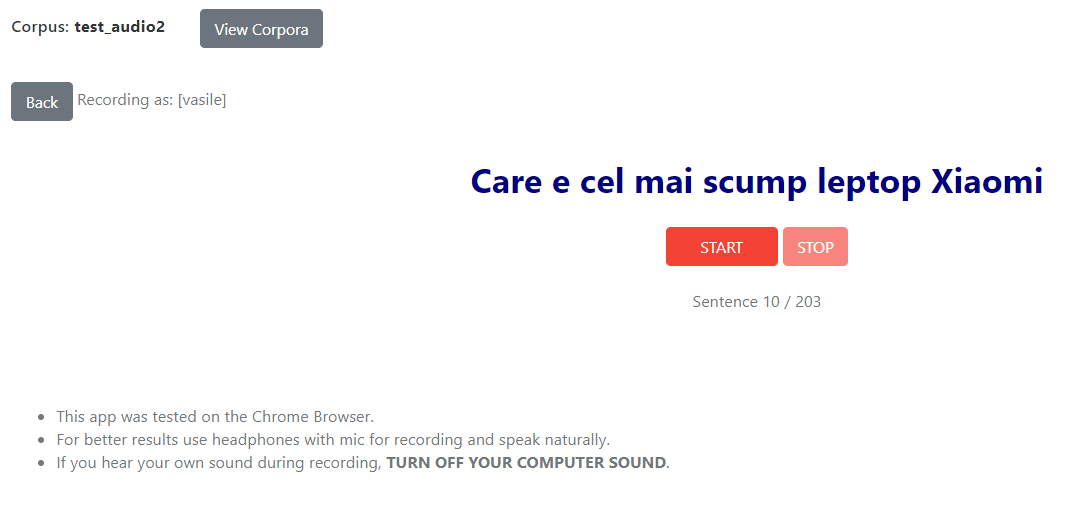}}
\caption{Sound recording component integrated in the RELATE platform.}
\label{fig:recording}
\end{figure}

Once the text files are uploaded, speakers can access the audio recorder. This is implemented using JavaScript and works within the RELATE general HTML template. When a speaker first accesses the component, it will ask for a pseudonym that will be used as part of the file name for all recordings. The speaker is presented with a single sentence and a "Start" button together with information about the current sentence number and the total number of sentences. Thus the speaker is offered the opportunity to first read and understand the sentence before starting the actual recording. The interface is presented in Fig. \ref{fig:recording}.

Recordings are stored as WAV files with a sample rate of 44.1 KHz using 16-bit signed integers. The recording component has a PHP back-end allowing it to store the files in the bimodal corpus, together with the associated text. In order to allow multiple speakers to record the same sentences, the file name incorporates the speaker pseudonym, thus creating unique file names for each of the speakers. Furthermore, in case of text uploaded as CSV files, the file name contains also the line number from the corresponding CSV file. 

We did not use a ``studio" environment for performing the recording. Instead, each speaker used his/her own hardware (headphones or dedicated microphone) to make the recordings. At any time after a sentence is recorded, the speaker (or another person given access to the corpus) can listen to the recording, download the associated WAV file and, if there were issues detected during recording (i.e., there was an unwanted noise or the speaker realises the pronunciation was not correct), delete it. The deletion of a recording will cause the associated sentence to re-appear in the recording component. This enables the speakers to re-record the sentences. 

After all the sentences were recorded, as part of the packaging process, the text was annotated using UDPipe\cite{b13} as integrated in the RELATE platform\cite{b14}. This provides linguistic annotations such as part-of-speech (using both universal part-of-speech tags\footnote{\url{https://universaldependencies.org/u/pos/}} and language-dependent MSD tags), lemmatization and dependency relations. No phonetic transcription is made. The resulting annotations are stored in tab-separated CoNLL-U\footnote{\url{https://universaldependencies.org/format.html}} files. 

Finally, a script was created to gather all the generated files (raw text, text annotations, sound recordings), anonymize the speakers, add metadata and create a single archive with the corpus. Text file names use the pattern \textit{Sn.txt} where \textit{n} is the sentence number (starting with 0 and ending with 710). Corresponding annotation files use the pattern \textit{Sn.conllu}. Sound files use the pattern \textit{Sn{\textunderscore}s.wav}, where \textit{n} continues to represent the sentence number and \textit{s} represents the speaker number (from 1 to 6). 

A metadata file was generated with corpus and speaker characteristics, including number of sentences, total duration, speaker's gender and age, number of recorded files by each speaker, information about recording device used. In order to anonymize the corpus, speaker's age is given only as intervals (for example "40-50" years).

\section{Corpus statistics}
\label{sec:statistics}
Statistics were computed at all levels: audio files, raw text and annotated text. For text-related statistics, the RELATE platform was used, while audio information was extracted using the \textit{soxi} utility from the \textit{SoX} (Sound eXchange)\footnote{\url{http://sox.sourceforge.net/}} software package. Audio statistics are given in Table \ref{tab:stats_audio} and text-related statistics are given in Table \ref{tab:stats_text}.

\begin{table}[h]
\caption{Audio statistics}
\begin{center}
\begin{tabular}{|l|c|}
\hline
\textbf{Statistic}&\textbf{Value} \\
\hline
Number of WAV files & 3786 \\ \hline
Total duration & 6h25m03s \\ \hline
Minimum duration & 1.02s \\ \hline
Maximum duration & 12.91s \\ \hline
Average duration & 6.10s \\ \hline
Total size & 1.89Gb \\ \hline
Sample rate & 44.1KHz \\ \hline
Channels & 1 \\ \hline
Encoding & Signed Int16 PCM \\ \hline
\end{tabular}
\label{tab:stats_audio}
\end{center}
\end{table}

\begin{table}[h]
\caption{Text statistics}
\begin{center}
\begin{tabular}{|l|c|}
\hline
\textbf{Statistic}&\textbf{Value} \\
\hline
Number of text files & 711 \\ \hline
Total text size & 57Kb \\ \hline
Maximum file size & 122b \\ \hline
Minimum file size & 3b \\ \hline
Average file size & 81.8b \\ \hline
Number of tokens & 11,927 \\ \hline
Unique tokens & 222 \\ \hline
Unique lemmas & 191 \\ \hline
Hapax legomena & 58 \\ \hline
\end{tabular}
\label{tab:stats_text}
\end{center}
\end{table}

The smallest text and the corresponding smallest audio recording, as indicated in the statistics tables, are associated with the simple interaction "Pa!" ("Bye!"). An example from an average sized text file is: ``Care e cel mai scump leptop acer, cu placă grafică dedicată tesla pe o sută și opt gigabaiți ram?" (``Which is the most expensive ACER laptop, with dedicated Tesla P100 graphical board and 8 gigabytes of RAM?"). The text in Romanian is written having in mind the pronunciation of English words and not their written form. Furthermore, numbers are written explicitly using words.

\begin{table}[h]
\caption{Most frequent 10 lemmas}
\begin{center}
\begin{tabular}{|c|c|}
\hline
\textbf{Lemma} & \textbf{Occurrences} \\
\hline
leptop & 605 \\ \hline
placă & 441 \\ \hline
grafic & 441 \\ \hline
gigabait & 419 \\ \hline
ram & 350 \\ \hline
dedica & 340 \\ \hline
scump & 220 \\ \hline
ieftin & 220\\ \hline
sută & 203 \\ \hline
mie & 200 \\ \hline

\end{tabular}
\label{tab:stats_lemma}
\end{center}
\end{table}

The most frequent lemmas (given in Table \ref{tab:stats_lemma}) show that most of the sentences are focused around the acquisition of laptops. Notice that the word for computer memory ("ram") appears in over half of the sentences. Even though the text corpus is rather small, the number of hapax legomena (words appearing only once) is rather reduced (only 58 words are hapax legomena as indicated in Table \ref{tab:stats_text}).

\begin{table}[h]
\caption{Most frequent 10 part-of-speech tags}
\begin{center}
\begin{tabular}{|c|c|c|}
\hline
\textbf{Tag}&\textbf{Occurrences}&\textbf{\# Unq. Lemmas} \\
\hline
NOUN & 2,675 & 66 \\ \hline
ADJ  & 1,698 & 32 \\ \hline
DET  & 1,211 &  7 \\ \hline
NUM  & 1,089 & 21 \\ \hline
ADP  &   919 &  9 \\ \hline
VERB &   558 & 29 \\ \hline
ADV  &   514 & 14 \\ \hline
PRON &   485 &  5 \\ \hline
AUX  &   467 &  3 \\ \hline
\end{tabular}
\label{tab:stats_pos}
\end{center}
\end{table}

The most frequent part-of-speech tags are presented in Table \ref{tab:stats_pos}. Nouns and adjectives are the most frequent ones. This answers the need of having computer parts with different characteristics covered by the corpus. Also the numerals are the fifth most frequent tags, corresponding to the different quantities associated with computer parts present in text.

The lexical diversity of the corpus is given by the number of unique lemmas and their proportion in the whole number of occurrences for each part of speech. As one can notice in the last column of Table \ref{tab:stats_pos}, the corpus is not very lexically diverse, as our aim was to capture a variety of ways in which relevant terms in this micro-world are pronounced.

\begin{table}[h]
\caption{Speaker statistics}
\begin{center}
\begin{tabular}{|c|c|c|c|}
\hline
\textbf{Spk}&\textbf{Gender}&\textbf{Age}&\textbf{Audio Files} \\
\hline
1 & M & 40-50 & 233 \\ \hline
2 & M & 30-40 & 711 \\ \hline
3 & M & 20-30 & 711 \\ \hline
4 & F & 30-40 & 711 \\ \hline
5 & F & 40-50 & 709 \\ \hline
6 & F & 40-50 & 711 \\ \hline

\end{tabular}
\label{tab:stats_speakers}
\end{center}
\end{table}

Speaker related statistics are presented in Table \ref{tab:stats_speakers}. This includes the gender, age group and the number of recordings.

\section{Corpus usage within the ROBIN project}
\label{sec:usage}

The primary reason behind the construction of the ROBINTASC corpus was the improvement of the ROBIN project's components involved in the micro-world scenario associated with a human-robot interaction in a computer store. The following sub-sections present an overview of the influence of this corpus on the software components: ASR and dialogue manager.

\subsection{Automatic Speech Recognition}
\label{sec:usage_asr}

The baseline ASR system \cite{b3,b4} was trained on 230 hours of Romanian speech and follows closely the Deep Speech 2 architecture \cite{b16}: 2 convolutional 2D layers \cite{b17}, 4 Long Short-Term Memory cells (LSTM) \cite{b18} of 768 neurons, 1 look-ahead layer \cite{b16} and 1 dense layer on top of which the softmax function was applied to create the output distribution over the possible characters. The ROBINTASC fine-tuned version of the baseline ASR system started with the baseline weights and completed the fine-tuning using the training part of the ROBINTASC corpus. The KenLM language model used to correct the transcriptions was also modified in the fine-tuned version of the ASR to better mimic the ROBINTASC words distribution, by multiplying each sentence 10 times in the text part of the training portion of ROBINTASC. The text replication step was performed in order to use an already existing automatic processing pipeline. This is not a limitation of the model itself which could have been adjusted using the model's weights instead of replicating the text.

The transcription performance was assessed on a test corpus that contains new sentences pronounced by one female and one male voice that also recorded samples in the ROBINTASC training part with the speaker id 5 (F5-test) and 1 (M1-test), respectively, together with a new male voice (M-new). It is known that WER (and CER) are better on sampled test corpora than on unseen data set \cite{b21}, containing voices that did not participate in the recording of the training data. Thus, we wanted to evaluate the close to real-world performance usage of the fine-tuned ASR system versus the baseline version.

The test corpus contains 50 questions that were designed to stress-test the ability of the fine-tuned ASR to adapt to the computer store domain. These sentences contain computer hardware-related companies that were found in ROBINTASC (e.g. Intel, CUDA, NVIDIA, etc.), but also new 
company names (e.g. Nokia, Siriux, etc.) or device names (e.g. "smart phone"). All English words have been phonetically transcribed to Romanian, following the design principles of ROBINTASC, in order to see if the ASR system can learn English pronunciations (e.g. "smart făun/fon" for "smart phone").

We evaluated both the baseline ASR system and the ROBINTASC fine-tuned ASR system on the test corpus. The results of the two versions are outlined in Table \ref{tab:asr_eval}. It can be observed that the fine-tuning process improved the performance of the model for all three speakers, improving the average WER by 16.3\% and the average CER by 7.8\%. The highest and the lowest improvements are obtained on the female voice from train (F5-test), 24.16\% WER, and on the male voice from train (M1-test), 10.34\% WER respectively. The performance on the new male voice (M-new) was enhanced by 14.33\% WER with fine-tuning.

\begin{table}[h]
    \caption{ASR evaluation results using baseline and fine-tuned versions on the three voices found in ROBINTASC test: male from train (M1-test), female from train (F5-test) and the new male voice (M-new).}
    \centering
    \begin{tabular}{|l|c|c|c|c|}
         \multicolumn{1}{c}{} & \multicolumn{2}{c}{\textbf{Baseline}} & \multicolumn{2}{c}{\textbf{Fine-tuned}} \\
         \hline
         \textbf{Model} & \textbf{WER} & \textbf{CER} & \textbf{WER} & \textbf{CER} \\ 
         \hline
         M1-test & 38.71 & 9.42 & 28.37 & 9.09 \\
         \hline
         F5-test & 81.23 & 48.41 & 57.07 & 29.71 \\
         \hline
         M-new & 59.21 & 26.28 & 44.88 & 21.83 \\
         \hline
         Average & 59.71 & 28.03 & 43.44 & 20.21 \\
         \hline
    \end{tabular}
    \label{tab:asr_eval}
\end{table}

Looking at the generated transcriptions, we can explain some of the errors in the following way:
\begin{enumerate}
    \item some clitics were not properly transcribed: "haș pe -ul" vs. "haș pe ul", "care-l" vs. "care -l" or "care îl" (see also the discussion on the treatment of clitics in \cite{SpedConsILR});
    \item one word is sometimes recognized as two consecutive words: "ultra portabil" instead of "ultraportabil";
    \item some of the English terminology in the test corpus has more than one possible phonetical transcriptions and in ROBINTASC all of these have been used: "uindous", "uindos" or "uindăus" for the English "Windows";
    \item in general, new English phonetically transcribed terminology is not properly recognized: "ol in oane" ("All in one") vs. "oli man" or "linăx cent ău es" ("Linux CentOS") vs. "linăx centes".
\end{enumerate}

Other reasons for the high WER values, for both the baseline and fine-tuned models, can be attributed to the different recording conditions and the amount of data used to train the models.

\subsection{Dialogue manager}
\label{sec:usage_dialogue}

The ROBIN Dialogue Manager (RDM, \cite{b20}) is a Java-based dialogue manager that works with micro-worlds. A micro-world is a set of definitions of spoken-about concepts, predicates that hold among them, ASR and TTS systems that work well in the micro-world and any other piece of information that would make an autonomous system (e.g. a robot) handle specific tasks in the micro-world. In the case of the notebook department of an electronics store micro-world, the robot should be able to give technical details and pricing for the existing stock of laptops.

RDM has been designed to work on the Pepper robot\footnote{\url{https://www.softbankrobotics.com/emea/en/pepper}}, enabling it to listen and respond to users' questions in Romanian. When enough information has been gathered through the conversation, RDM can supply predefined action items to the robot's planning algorithm., e.g. ``Let me find out if your laptop is in stock."

We have empirically evaluated RDM with the baseline and fine-tuned ASR systems, by asking it different questions, appropriate to the electronics store micro-world. While we do not have a quantitative evaluation on how much better the fine-tuned ASR system is, it was significantly better than the baseline ASR system mainly because English terminology was not handled \emph{at all} by the baseline model but was handled acceptably well by the fine-tuned ASR system, as long as the English terms were in the ROBINTASC corpus, e.g. "leptop" (English "laptop"), "haș pe" ("HP"), "gigabaiți" ("GB"), "epăl" ("Apple"), etc. This is an indication that the fine-tuned ASR can be further improved with new English terms, should the need arise.

\section{Conclusions}
\label{sec:conclusions}
This paper introduced ROBINTASC, a new Romanian language speech corpus from the ROBIN project. We have shown that it had a positive influence on two components developed within the ROBIN project, namely an ASR system based on Deep Speech 2 architecture and a dialogue manager, developed for micro-world scenarios. The corpus is open sourced, available under a Creative Commons Attribution NonCommercial NoDerivatives (CC BY-NC-ND) 4.0 license\footnote{\url{https://creativecommons.org/licenses/by-nc-nd/4.0/}} and can be downloaded from the Zenodo platform\footnote{\url{https://doi.org/10.5281/zenodo.4626540}}.

\section*{Acknowledgment}

The research described in this article was supported by a grant of the Romanian National Authority for Scientific Research and Innovation, CNCS – UEFISCDI, project number PN-III 72PCCDI ⁄ 2018, ROBIN – “Roboții și Societatea: Sisteme Cognitive pentru Roboți Personali și Vehicule Autonome”.

\end{document}